# Applying Quantum Hardware to non-Scientific Problems: Grover's Algorithm and Rule-based Algorithmic Music Composition

Alexis J. Kirke, Senior Research Fellow, Interdisciplinary Centre for Computer Music Research, University of Plymouth, Drake Circus, Plymouth PL4 8AA, UK

**Abstract:** Of all novel computing methods, quantum computation (QC) is currently the most likely to move from the realm of the unconventional into the conventional. As a result some initial work has been done on applications of QC outside of science: for example music. The small amount of arts research done in hardware or with actual physical systems has not utilized any of the advantages of quantum computation (QC): the main advantage being the potential speed increase of quantum algorithms. This paper introduces a way of utilizing Grover's algorithm - which has been shown to provide a quadratic speed-up over its classical equivalent - in algorithmic rule-based music composition. The system introduced - qgMuse - is simple but scalable. Example melodies are composed using qgMuse using the ibmqx4 quantum hardware. The paper concludes with discussion on how such an approach can grow with the improvement of quantum computer hardware and software.

## 1. INTRODUCTION

Why – out of all of the non-conventional methods of computation - are quantum computers (QC) attracting so much government and private funding? The answer is speed [1, 2]. Quantum mechanics (QM) models the world by considering a physical state as a sum of all its possible configurations. For example, the physical state of an electron is modeled as a weighted sum of a large number of vectors (called eigenvectors), each of which represents a possible result of a measurement in the laboratory. This sum of vectors – called a superposition - varies over time as the electron's state changes. When the electron's state is measured at a particular time in the lab, the result will be based on just one of the vectors. This natural parallelism of superposition together with a property known as entanglement (the ability of entirely separated particles to effect each other instantly), are both candidates for explaining the cause of the speed-up - compared to classical computers - found in certain quantum algorithms [3]. Another candidate is 'interference'. The wave-particle nature of states in quantum computing will not be discussed in depth here, as it is not helpful to the explication of the main points of this paper. But suffice it to say that states in QM can be represented as complex waves. These waves might interfere in ways that cause the calculation to focus more rapidly on the desired result. Quantum computers are able to utilize these elements because they represent data using physical quantum objects. For example, by confining ions using radio-frequency electric fields [4], a superposition of data can be set up and manipulated in a controlled way.

Because of differences between quantum and classical physics, certain quantum algorithms have been theoretically proven to be orders of magnitude faster than their classical equivalents. (Note that although researchers are not agreed on the precise cause of the quantum speed-up, it has still been possible to prove mathematically that the computational speed-up occurs.) Shor’s algorithm [1] is exponentially faster at breaking public key encryption than the fastest non-quantum algorithm. This is seen as a serious potential security threat [5]. Grover’s algorithm [2] is quadratically faster than the best classical algorithm at performing an unstructured-database search or function inversion.

Another feature of QM is its probabilistic nature. The results of measurements on an electron in general cannot be predicted with certainty. QM simply provides a means to calculate the probability of the electron being in a certain state. Surprising results relate to this. In classical physics if an electron is fired at a sufficiently strong electromagnetic barrier, it will fail to penetrate it. In quantum mechanics, there is a probability that the electron will be observed on the other side of the barrier. This is known as quantum tunneling. In quantum computing (QC), this tunnelling becomes relevant when building quantum annealers [6]. Annealers can be thought of as traversing a fitness landscape looking for the global minimum. One main weakness is that the solver can get trapped in a local valley - i.e. it thinks it's at the bottom of a valley, but in fact just over the hill is a much deeper valley. However the solver can't "see" it, because of the hill. In the quantum version of this algorithm, the quantum solver can tunnel through the mountain to the lower valley, leading to potential speed-ups in solving [7].

As an aside: the non-deterministic nature of QC has an interesting conceptual implication for algorithmic artists. Artistic algorithms have utilized pseudo-random algorithms since the first computer arts were created, right up to some of the most recent creations. This is because pseudo-randomness helps to prevent deterministic algorithms from producing overly repetitive output. Pseudo-random numbers are not truly random [8] (because classical computers are deterministic) and their sequence will repeat eventually. QC is not pseudo-random. The most prevalent interpretation of QC amongst researchers is that it is non-deterministic and has randomness at its heart. A quantum algorithm for which there is a desired deterministic result needs to be run multiple times to get a statistically significant final output. The final output is some averaging of all the intermediate outputs. Such a form of computation provides a new way of thinking about randomness in computer arts. Rather than trying to create randomness from determinism - as in classical computing, QC requires determinism and complexity to be built from randomness. The implications of this reversal of thinking for the arts are hard to imagine at this stage. These questions can only be answered by starting to apply basic quantum algorithms to the arts.

As already mentioned, two main potentially useful algorithms – Shor and Grover - have been identified, but neither has been implemented in hardware in a way that utilizes their proven quantum speed-up. Developing quantum algorithms requires a new way of thinking: rotations in complex vector spaces, probabilistic results, entanglement and superposition. Programming quantum computer algorithms requires an undergraduate level of mathematics and physics. The author has discussed this issue in greater depth here [9]. The purpose of this paper is to develop and test a computer music algorithm qgMuse, that is implementable on a hardware quantum computer, and that does utilize QC as a solution – i.e. it aims to use QC to do things in a way a non-QC could not do.

The reality is that no quantum computer exists that can deal with useful versions of the quantum algorithms mentioned above. For example, the most powerful QC algorithm - Shor's Algorithm for factoring into primes - has been used on a hardware computer recently to factor 15 into 3 and 5 [10]. This is a factorization that is trivial can clearly be done by hand. The only QCs which claim to be ready for commercial work are the quantum annealers mentioned earlier – made by D-Wave [6]. These quantum annealers cannot run Shor’s or Grover’s algorithm – the future “killer apps” of QC. The type of QC that can run Shor's and Grover's algorithms are called "gate-based" quantum computers. As will be seen later, gate-based QC algorithms such as Shor and Grover can be visualized in a way reminiscent of that used to visualise classical logic-gate circuits.

Faced with this, a computer arts researcher may be tempted to simply drop all investigation into gate-based quantum computers - those that can run Shor and Grover - and focus only on quantum annealers, until more powerful and stable gate-based quantum computers are available. This view seems ignorant of – for example - musical history. Computer music began its research with simple bleeped tunes on early mainframes, and developed in parallel with the development of computing. This paper argues for a similar approach for gate-based quantum computer music. Much previous research in gate-based QC has been done in simulation or theory. There are a couple of exceptions - for example [11, 12] - but none attempted to utilize the quantum algorithms known to give definite and large quantum speed-ups. Work needs to be done on actual quantum computers in these early stages, to ensure that computer arts keeps pace with advances in quantum computing, and also to see exactly what is feasible on a quantum computer in artistic and musical terms. So although the music generated by the research reported here is very simple, the paper’s contribution lies not in the quality of that music but in the foundations that the present work lays for future artistic work as quantum computing matures into full “conventionality”.

## 2. RELATED QUANTUM MUSIC WORK

Previous designs for performances and music involving quantum mechanical processes have either been metaphorical [13], based on simulations (online [14] or offline [15]), not utilized the quantum speed-up [11,16], or - in the case of actual real-time physics performances - not been directly concerned with quantum effects [17,18]. It is important to take a moment to define what is meant in this paper by "utilizing the quantum speed-up". There are no algorithms on gate-based quantum computers available that perform tasks faster than a classical computer. The killer quantum computer algorithms have been proved to be faster only in theory. The speed increases, however, are so vast that this has led to the large amount of money being poured into quantum technology research. Furthermore, the implementations of the algorithms on current gate-based quantum computers are at a level where the problems they solve are trivial - for example searching a database of sixteen 1-bit entries, or showing that 15 can be written as 3 times 5. However these implementations are theoretically scalable. Thus when this paper refers to a system "utilizing the quantum speed-up", it means: (a) that the system is based on a quantum algorithm that has been proved to be theoretically much faster than its classical counterpart, and (b) that the system is in theory scalable so that even if it is a trivial example, it could eventually incorporate examples that will run faster than their classical counterparts.

In terms of offline simulations, one of relevance to this paper is the web page Listen to the Quantum Computer Music [19]. Two pieces of music are playable online through MIDI simulations. Each is a sonification of the two key quantum computation algorithms: Shor's and Grover's. The offline sonification of quantum mechanics equations have also been investigated in [20,21] and [22], with the third sonifying Large Hadron Collider data from The European Organization for Nuclear Research (CERN) to create a musical signature for the (at-the-time) undiscovered Higgs Boson. Another paper defines what it calls Quantum Music [13] in simple theory form, though once again this is by analogy to the equations of quantum mechanics, rather than directly concerned with quantum computing. It examines what one might call the “trivial” representation in quantum music. Each note in a melody is a superposition of all possible notes. It has not been implemented on a hardware QC. It would need to be mapped into the QC realm and then into the hardware QC realm. After that, working with, say, two melodies that are superpositions of 8 notes each – for example entangling them - would require significant circuit complexity. Current hardware quantum computers would not be able to cope with them, purely from a stability point of view (as will be seen later). Most importantly, even if it could be implemented, the presented formalism does not utilize the quantum speed-up.

Certain equations of quantum mechanics have also been used to synthesize new sounds in simulation [15]. The orchestral piece “Music of the Quantum” [23] was written as an outreach tool for a physics research group, and has been performed multiple times. The melody is carried between violin and accordion. The aim of this was as a metaphor for the wave particle duality of quantum mechanics, using two contrasting instruments.

The most impressive quantum simulation performance has been Danceroom Spectroscopy [14] in which quantum molecular models generate live visuals. Dancers are tracked by camera and their movements treated as the movement of active particles in the real-time molecular model. Thus the dancers act as a mathematically accurate force field on the particles, and these results are seen in large scale animations around the dancers.

There have been performances and music that use real-world quantum-related data.

However most of these have been done offline (not using physics occurring during the performance). These include the piece Background Count: a pre-recorded electroacoustic composition that incorporates historical Geiger counter data into its creation [24]. Another sonification of real physics data, but done offline, was the LHChamber Music project [25], instrumented for a harp, a guitar, two violins, a keyboard, a clarinet and a flute. Different instruments played data from different experiments (for example ATLAS and LHCb).

The first real-time use of subatomic physical processes for a public performance was Cloud Chamber [17]. In Cloud Chamber physical cosmic rays are made visible in real-time, and some of them are tracked by visual recognition and turned in to sound. A violin plays along with this, and in some versions of the performance, the violin loudness level triggered a continuous proportional electric voltage that changed the subatomic particle tracks, and thus the sounds (creating a form of duet). Cloud Chamber was followed a few years later by a CERN-created system which worked directly, without the need to use a camera. Called the Cosmic Piano, it detects cosmic rays using metal plates and turns them into sound [18]. However it had no feedback loop from the acoustic instrument to the cosmic ray tracks, unlike Cloud Chamber.

The previous two discussed performances were live, and the data was not quantum as such. It was quantum-related in that the cosmic rays and cloud chambers are subatomic quantum processes. But the performances do not incorporate actual controlled quantum dynamics or computation in their music. [26] created sound with quantum computing but was primarily about connecting other forms of unconventional computing (PMAP) to a quantum computer. It was designed for use with an online photonic quantum computer, however for technical reasons the computer was taken offline, and so the final results were generated using the online simulator. The paper included the use of the system to compose an orchestral piece of music that musified a photon-based quantum gate called a CNOT, approaching maximum entanglement.

The first use of controlled quantum dynamics in hardware quantum computation to make music was the algorithm qHarmony [6]. It was implemented on an adiabatic quantum computer and also utilized in real-time in a live music performance with a mezzo-soprano [27]. The first use of gate-based quantum computer hardware to make music was the algorithm GATEMEL [11,12]. This algorithm does not utilize the quantum speed-up but only the non-deterministic nature of QC. A second algorithm tested on a gate-based hardware QC that utilizes the non-determinism is Quantum Music Composer [16], which is a step on from GATEMEL in that it implements a Markov chain and harmonies, but it, too, does not utilize the quantum speed-up.

## 3. RULE-BASED ALGORITHMIC MUSIC COMPOSITION

The quantum algorithm introduced in this article is utilized to support rule-based algorithmic composition. The use of rule-based or knowledge-based methods for algorithmic composition have been common for many years [28]. In such an approach, the composer/user predefines a set of rules that can generate or constrain musical features. One of the first algorithmic compositions, the Illiac Suite [29] involved randomly generating notes and then dropping notes which did not fit the rules of certain composition styles – for example textbook counterpoint for the second movement. Since then rule-based systems have developed where the rules can be used to partially or fully generate the actual music features.

Rules can be applied in a bottom-up or top-down approach. For the bottom-up approach the rules are the generative engine themselves. For example, let r(t) be a function that generates a pseudo-random non-negative integer at time t. Then a rule to generate an even numbered musical feature at time t would be F(t) = 2r(t). Or rules to generate two MIDI pitches p(t) and p*(t) with 5 semi-tones difference would be:

$$p(t) = r(t) \bmod 12 + 60 \tag{1}$$

$$p^*(t) = p(t) + 5 \tag{2}$$

Now consider the top-down approach. In these cases a feature is generated - usually pseudo-randomly - and then checked against the rule. For example, to implement the rule F(t) = 2r(t) as a top-down rule, a random number R can be generated. Then it can be factored to examine if there exists an integer n such that R = 2n to fulfill the rule. Or for an intervallic example, if two pitches p(t) and p*(t) are randomly generated, it can then be checked if they fulfil the rule:

$$|p(t) - p^*(t)| = 5 \tag{3}$$

This article is inspired by the top-down rule approach. Such rules can be highly contextual (i.e., melodic or harmonic intervals) as well – for example the allowed pitch distance between adjacent or coincident notes could be constrained based on the previous 5 distances looking back in time. Rules can cross reference each other – the allowed adjacent pitch distance could be limited by the allowed coincident distance. The compositional style in such a top-down rule-based system usually comes not only from the individual rules, but how they are logically combined. For example, suppose two sub-rules are defined for randomly generated pitches:

$$|p(t) - p(t-1)| > 0 \tag{4}$$

$$p(t) - p(t)^{*} > 1 \tag{5}$$

Two of the possible methods of combining these sub-rules to make a rule could be:

$$(|p(t) - p(t-1)| > 0) \cdot (p(t) - p(t)^{*} > 1) = 1 \tag{6}$$

$$(|p(t) - p(t-1)| > 0) \oplus (p(t) - p(t)^{*} > 1) = 1 \tag{7}$$

The first version ANDs the sub-rules - it only gives value 1 if both sub-rules are satisfied. The second version XORs the sub-rules - it only gives value 1 if only one of the sub-rules is satisfied. These are clearly going to give significantly different musical outcomes. It is this Boolean approach to rule specification that is usually taken in top-down rule-based systems. A significant part of the generative process in such systems is solving equations such as (6) and (7) to check if the features satisfy the equations. However a simple generate-and-test approach is generally considered inefficient.

When the number of sub-rules and their combinations grows - as is required for musically interesting and relevant systems - the generate-and-test approach slows exponentially. For example, the groundbreaking CHORAL system [30] used 350 rules in a Boolean-type form. These rules are designed to capture the style of J. S. Bach for four-part harmonies. Once these levels of complexity are reached, simple generate-and-test is unfeasible, and methods such as constraint programming and backtracking are used [31] to speed up the search for musical solutions. At this point the rules become labeled as constraints. The musical problem in general is in fact so complex, that musical constraint programming has at times helped to drive research in general constraint programming. Because the search space in such problems is often huge, an efficient solver has a large impact on the usability of a rule-based system [32]. As recently as 10 years ago, the state of the art in most musical constraint systems meant that problems for polyphonic music in which both pitch structure and rhythmical structure were constrained (as opposed to one or the other), were considered hard or even impossible to solve [33]. Applying the wrong strategies in complex rule-based composition systems can slow down the search speed by orders of magnitude [34]. Furthermore, new themes have been emerging in the last 10 years, such as real-time constraint-based systems [35, 36] which are even more speed sensitive. For example [36] sets a time-out on finding a solution, due to the risk of it taking too long to find one in real-time operation.

Obviously many of these problems will be addressed through research into more efficient constraint-based algorithms. However, there is a key assumption behind the use of techniques such as constraint programming in rule-based algorithmic composition, that makes quantum computing relevant. This assumption is the inefficient speed of unstructured random search. This issue is reminiscent of a similar assumption made early in the evolution of chess computers. In Shannon's seminal 1950s paper on computer chess [37], it was stated that computer chess machines would not be able to perform an exhaustive ("brute-force") search of future game possibilities due to such a search taking too long. Thus a number of heuristics and alternative methods were developed and used in chess-playing algorithms.

However as classical computation speeds increased beyond what Shannon could have imagined, brute-force searches played a much greater role in chess computer strategies than Shannon originally envisioned [38]. Modern chess computers still use heuristics, but brute-force is a key part of their strategies as well. Analogously, if quantum computers can be built that can run unstructured random searches accurately and efficiently, it is entirely feasible that the speeds attained by quantum search algorithms will reduce the need for, or augment, some of the approximations, algorithms and heuristics used in solving rule-based composition systems. This can lead to combinations of the fast quantum "brute" search with the best classical rule-based solvers.

Before detailing such a quantum search algorithm, another example of unstructured search will be considered. Boolean equations (6) and (7) are extremely simple, and can be solved quickly by eye. In general a Boolean equation of sub-rules $sr_i$ could be more complex, such as:

$$\big((sr_1 \cdot sr_2 \cdot sr_3{}') \oplus (sr_2{}' + sr_4) \cdot sr_3\big) + \big((sr_1 \cdot sr_5{}' \cdot sr_6) \oplus (sr_6 + sr_8)\big) + \big((sr_9 \cdot sr_{10}{}' \cdot sr_{11}) \oplus (sr_{10} + sr_{12})\big) = 1 \qquad (8)$$

(Note that the apostrophe ' is a logical NOT operation). To find the truth values of the $sr_i$ by eye is not simple. Solving equation (8) using unstructured random search means all possible values of the $sr_i$ are tried (0 or 1) until the solutions are found. This takes, in the worst case, $2^{12} = 4096$ iterations. Easily do-able on a fairly old desktop in a trivial amount of time. But as the number of sub-rules grows, the number of average iterations needed to find a solution grows rapidly. If there are 30 sub-rules, then the number of iterations required to find the solution is, on average, $(2^{30})/2 = 536{,}870{,}912$. When multiplying a matrix, an i5 Intel chip performs roughly 65536 multiplication operations per millisecond [39]. So, conservatively, assume each calculation of an iteration in the 30 sub-rules takes a 100th of a millisecond, that would be about 1.5 hours to generate one time-step of the music features in the rule-based system. As already mentioned, such a naive approach is not usually used.

However, consider a search algorithm that is quadratically faster at unstructured random search – i.e. rather than taking N iterations it took $\sqrt{N}$ searches. For a system that is quadratically faster, the 30 sub-rule example above would take under 4 milliseconds. If there were 40 sub-rules, a quadratic speed up would be relatively more dramatic. Such a quadratically faster search has been proven to exist and can be implemented on quantum computer, currently for a much smaller number of sub-rules, but in a scalable way. The purpose of this paper is to introduce a small sub-rule system in a scalable way, and implement it on quantum hardware.

The fast search is the already-mentioned Grover's algorithm [5] and requires that the Boolean function can be implemented on a quantum computer. The fact that Grover's algorithm is implemented on a quantum computer will lead to it being both generative and soft - which will be seen to be useful features of the system introduced later in this paper. Each time Grover is run and collapsed, it generates a suggested solution to the Boolean function, with those that satisfy the rule having the highest probability of being generated. Suppose there are, say, 1,000 correct solutions to a combined rule, and 1 billion incorrect solutions. Then Grover will - most of the time - randomly select one of the correct solutions, but with a lower probability it may output one of the incorrect values. This type of feature is desirable in soft rules, and such soft rules are sometimes desirable in computer music systems [31].

It is worth mentioning that, in fact, constraint-based composition systems are currently extremely efficient in terms of speed for a large subset of musical problems. However, consider a future target outside of this subset: for example composing a 70 minute choral symphony complete with instrumentation. Such a rule-based system (real-time or otherwise) would have to wait a long while until Moore's Law allows it to run in practical time scales. But there is no guarantee that Moore's Law will continue to apply for many years ahead [40]. Hence there is no certainty that classical computers will reach the necessary speed to create such a rule-based symphony.

## 4. GROVER'S ALGORITHM

Grover's algorithm can now be introduced using the Dirac notation. The simplest practical introduction to Grover's algorithm currently available is probably [41]. At the heart of Grover's algorithm is a representation of the Boolean function to be inverted – called the Oracle. Usually the Oracle designed around a quantum form of the function to be solved. In the case of rule-based music composition, it represents the rule to be solved. The first step is to define the rule inputs as qubits. So if the rule has three binary inputs, then the classical version could be written:

$$\mathrm{r}(\mathrm{i}_0, \mathrm{i}_1, \mathrm{i}_2) = 1 \tag{15}$$

The quantum version is written:

$$R|i_0 i_1 i_2\rangle \text{ or } R|q_0 q_1 q_2\rangle \tag{16}$$

since the inputs will have to be represented as qubits and the rule as a quantum operator R. In the algorithm, the inputs [$i_0$, $i_1$, $i_2$] are initialized to 0-valued qubits, giving

$$|q_0 q_1 q_2\rangle = |000\rangle \tag{17}$$

Next in the Grover algorithm, the qubits are put into a superposition of all their possible values of [$i_0$, $i_1$, $i_2$] (these are 000, 001, 010, 011, 100, 101, 110, 111). This is done using the three "parallel" H gates. Applying them across the 3 qubit state in (17) (and ignoring the scalar multipliers for brevity) will give:

$$A = H^{\otimes 3}|000\rangle = |000\rangle + |001\rangle + |010\rangle + |011\rangle + |100\rangle + |101\rangle + \cdots \tag{18}$$

The notation $H^{\otimes 3}$ represents the fact the H is acting on a three qubit state. Next the Oracle operation R is performed on this superposition A. The Oracle must be represented as a unitary matrix that multiplies the superposition:

$$B = R(A) = R(|000\rangle + |001\rangle + |010\rangle + |011\rangle + |100\rangle + |101\rangle + \cdots) \tag{19}$$

$$B = R|000\rangle + R|001\rangle + R|010\rangle + R|011\rangle + R|100\rangle + R|101\rangle + \cdots \tag{20}$$

The Oracle has one key function: it must flip the phase of the part of the superposition A in equation (18) that represents the correct solution to the rule - i.e. for which $r(i_0, i_1, i_2) = 1$. This can be done by calculating a quantum version of r in such a way that the state for which r is satisfied has a negative phase. Suppose the correct solution to $r(i_0, i_1, i_2) = 1$ is 0,1,0. Then applying R in equation (20) should give:

$$B = |000\rangle + |001\rangle - |\mathbf{010}\rangle + |011\rangle + |100\rangle + |101\rangle + \cdots \qquad (21)$$

Next comes the key part. A quantum function is applied, that moves the superposition of ($i_0$, $i_1$, $i_2$) towards a value that, if measured, now has a higher probability of giving $r(i_0, i_1, i_2) = 1$ (the desired result). The operator that does this is called the Grover diffusion operator, and is (in the 3 qubit case):

$$R_0 = H^{\otimes 3}\,(2|000\rangle\langle 000| - I)H^{\otimes 3} \qquad (22)$$

$\langle 000|$ is the complex conjugate vector of $|000\rangle$. I is the identity matrix - it is a quantum function which when acting on a vector, simply returns the same vector. Applying the $R_0$ operator to the superposition B in equation (20) has one key effect. It inverts the weightings of B around their mean. The mean of the weightings will have been reduced by making the solution states have negative weightings, so inverting all weightings around the mean will increase the size of the negative weighted items. These are the items which are solutions to equation (15). The combination $R_0R(A)$ can also be thought of as a rotation of the superposition in a complex vector space in such a way as to move the superposition vector A of equation (18) towards the solution. These operations amplify the probability of observing the ($i_0$, $i_1$, $i_2$) which satisfy equation (15). This amplification is very rapid. After less than three applications of Grover (in the 3 input rule case), if the quantum state of the Grover output is observed, then the most likely result will be the values ($i_0$, $i_1$, $i_2$) that give $r(i_0, i_1, i_2) = 1$.

The fact that quantum algorithms are probabilistic raises another issue. After less than 3 iterations of a 3 qubit input Grover's algorithm, although the most likely observation will be values of ($i_0$, $i_1$, $i_2$) that obey (15), there is a non-zero probability of measuring the wrong values. This means that if r is meant to be a "hard" rule, a quantum algorithm needs to be sampled multiple times in order to get the result required. However, it is an axiom of QM that observing the output of a quantum algorithm leads to a collapse of the superposition. Thus the algorithm needs to be run again to get another sample output. This can - of course - cancel out the quantum speed up for hard rules. For example, using only 3 inputs as above, the quantum algorithm might have to be run 20 times on the ibmqx4 to give confidence in the result. Then in real terms 40 iterations have been performed, rather than the up-to-8 iterations with the classical version. However this disadvantage quickly disappears as the number of inputs increases, given the speed-up factors demonstrated earlier. Also, if a soft rule is desired, then the user may be happy to select the first output given.

## 5. IMPLEMENTING RULES ON A HARDWARE QC

The advantages of Grover's algorithm were introduced earlier in this paper in the context of larger-scale rule based system such as CHORAL; i.e. a system that uses constraint programming or advanced Boolean solving tools to deal with the larger numbers of rules and inputs. Using a Grover of this scale would be too complex to provide an introductory example. Hence a small-scale Grover is utilized in this paper to build the foundations of scalable approach.

### 5.1 qgMuse

qgMuse is a scalable hybrid hardware classical/quantum algorithm that was implemented using an IBM quantum computer - the ibmqx4. The IBM quantum computer is housed in a large dilution refrigerator, supported by multiple racks of electronic pulse-generating equipment. The qubits used are known as fixed-frequency superconducting transmon qubits, and are Josephson-junction based to reduce noise effects [42]. They use fixed-frequency qubits to minimize sensitivity to external magnetic field fluctuations that could corrupt the quantum information. The superconducting qubits are made on silicon wafers with superconducting metals such as aluminium and niobium. The processor is contained in a printed circuit board package shielded within a light-tight, magnetic-field shielding can. The dilution refrigerator cools the device down to around 15 milliKelvin. It works by circulating a mixture of two Helium isotopes. Electromagnetic impulses at microwave frequencies are sent to the qubits through coaxial cables with a particular phase, duration, and frequency. These enact the quantum gates.

To measure the qubits, each is coupled to a microwave resonator. A microwave tone is sent to the resonators, and the qubits state can be retrieved from the phase and amplitude of this reflected signal. Signals in the resonator are amplified within the dilution refrigeration layers: a quantum-limited amplifier at 15 mK, and a high-electron mobility transistor amplifier at 4K. The system is re-tuned three times a day, which takes up to an hour. There are many opportunities for the qubits and their entanglements and superpositions to become corrupted by external temperature and electromagnetic interference and other elements.

qgMuse implements two rules, partly inspired by the Narmour Implication-Realisation (I-R) model of melodic expectation [43]. It is designed to demonstrate a scalable quantum model of composition that will be able to utilize practical speedups from Grover's algorithm once they become available. It is done in the spirit of Monz et al. [10] which used Shor's quantum algorithm to factor 15 into 5 and 3 on quantum hardware - as a vital and scalable link in quantum computer music research. The fact Shor's algorithm can be used in a simplified form on current QC hardware, is a major step towards using it in situations where the quantum speed-up becomes relevant. Similarly with qgMuse: although it is implemented in a way that doesn't require a quantum speed-up, it lays the framework for a future computer music system based around the same algorithm, that does.

The Narmour I-R model is a proposed approach utilizing soft rules for the behaviour of sequences of musical notes. One of the I-R "rules", and one of its assumptions will be interpreted into soft Boolean rules. Firstly the I-R "Registral direction" rule, which states that large pitch intervals tend to precede a pitch direction change, whereas small intervals tend to be continued in the same direction. A large interval will be defined here as spanning 4 white piano notes or more. So the pitch jump c -> f is large, whereas c -> e is small. Or f -> b is large, whereas f -> a is small. qgMuse only uses white notes here as it can only have a limited number of rules for practical quantum implementation. Hence the tonality is restricted at the outset, rather than implementing it on the QC hardware as a rule. The large interval flag $\mathrm{LI}(\mathrm{t})$ will be defined as 1 if the number of white notes from the current pitch to the previous pitch at $\mathrm{t}-1$, including both end pitches, is 4 or more, and $\mathrm{LI}(\mathrm{t})$ is 0 otherwise. Define the direction change flag $\mathrm{DC}(\mathrm{t})$ as 1 if the melodic interval from $t{-}1$ to $t$ is in the opposite direction to the interval from $t-2$ to $t-1$, and 0 otherwise. Then consider the XOR-based rule:

$$(\mathrm{LI}(\mathrm{t}-1) \oplus \mathrm{DC}(\mathrm{t}))' = 1 \tag{23}$$

This is satisfied if $\mathrm{LI}(\mathrm{t}-1)$ and $\mathrm{DC}(\mathrm{t})$ are equal, i.e. if a large interval is followed by a direction change, or if a small interval is followed by no direction change. So if this is implemented as a fuzzy or soft rule, it is compatible with the "Registral Direction" tendencies. The following rule will also be softly implemented:

$$\mathrm{LI}(\mathrm{t}-1)' \cdot \mathrm{LI}(\mathrm{t})' = 1 \tag{24}$$

which says that all intervals are small (non-large) intervals following small (non-large) intervals. Such a rule would not be useful in a hard-rule generative music system, but is useable in a soft-rule system to make small intervals dominate over large ones - an assumption in Narmour's I-R approach. The two rules between them use 3 variables, which requires a 3-input Grover to solve equation (25), made up from the two sub-rules above:

$$(\mathrm{LI}(\mathrm{t}-1) \oplus \mathrm{DC}(\mathrm{t}))' \cdot \mathrm{LI}(\mathrm{t}-1)' \cdot \mathrm{LI}(\mathrm{t})' = 1 \tag{25}$$

The quantum part of qgMuse is shown in Figure 1. Figure 1's circuit solves the rule $(\mathrm{LI}(\mathrm{t}-1) \oplus \mathrm{DC}(\mathrm{t}))' \cdot \mathrm{LI}(\mathrm{t}-1)' \cdot \mathrm{LI}(\mathrm{t})' = 1$ using Grover's algorithm. The oracle is marked up with three rounded rectangles numbered (1), (2) and (3). Everything after the markups is part of the Diffusion operation, up until the final three boxes/gates, which measurement the outputs.

$\mathrm{LI(t)}$, $\mathrm{DC(t)}$, and $\mathrm{LI(t-1)}$ are represented by $q_0$, $q_1$ and $q_2$ respectively. The qubit "inputs" are all initialized to $|0\rangle$. Rectangle (1) uses a CNOT gate to take the XOR of $q_2$ and $q_1$, represented on $q_1$. It then uses an X gate to take the NOT of $q_1$. Thus $q_1$ now represents $(\mathrm{LI(t-1)}\ \oplus \mathrm{DC(t)})'$. The X gates on $q_0$ and $q_2$ perform a NOT on $q_0$ and $q_2$ to represent $\mathrm{LI(t-1)}'$ and $\mathrm{LI(t)}'$ respectively. Rectangle (2) implements a control-control-Z (ccz) gate, specifically $\mathrm{ccz}(q_2, q_1, q_0)$. This can be seen by multiplying out the matrices of the operations, but will not be demonstrated here for space reasons.

Given that $q_1$ represents $(\mathrm{LI(t-1)}\ \oplus \mathrm{DC(t)})'$, $q_2$ represents $\mathrm{LI(t-1)}'$ and $q_0$ represents $\mathrm{LI(t)}'$ then - by the definition of control-control-not $\mathrm{ccz}(|q_0q_1q_2\rangle)$ - $|q_0q_1q_2\rangle$ has its phase flipped if $q_0 = 1$ and $q_1 = 1$ and $q_2 = 1$. In other words, if $(\mathrm{LI(t-1)}\ \oplus \mathrm{DC(t)})' = 1$ and $\mathrm{LI(t-1)}' = 1$ and $\mathrm{LI(t)}' = 1$. Thus, as required, the Oracle represents equation (25). Rectangle (3) simply reverses the operations performed in Rectangle (1) before going through the Diffusion operations. The Diffusion operations (which need $q_0$ and $q_1$ to have the same values that had before rectangle (1)) represent $R_0$ in equation (22). When this circuit has its output observed, it should provide, with the highest probability, solutions to equation (25).

The non-quantum part of the qgMuse works as follows. It starts on a base note – middle C will be used - at time step $t = 0$. Then it runs the quantum circuit to invert equation (25) - to get the allowed values of $\mathrm{LI(t-1)}$, $\mathrm{DC(t)}$ and $\mathrm{LI(t)}$. Backtracking is not utilized in qgMuse, so if the allowed $\mathrm{LI(t-1)}$ from the QC is different to the actual $\mathrm{LI(t-1)}$ from the previous time step, then the whole generation process is skipped and new candidates are requested from the QC. In other words if the previous white note interval was large but the quantum algorithm returns a solution involving the previous interval being small - or vice versa - then qgMuse calls the quantum part of the algorithm again. If the rules were hard rules, then the returned solution would not change on this second call. But for the non-deterministic Grover, the solutions can change.

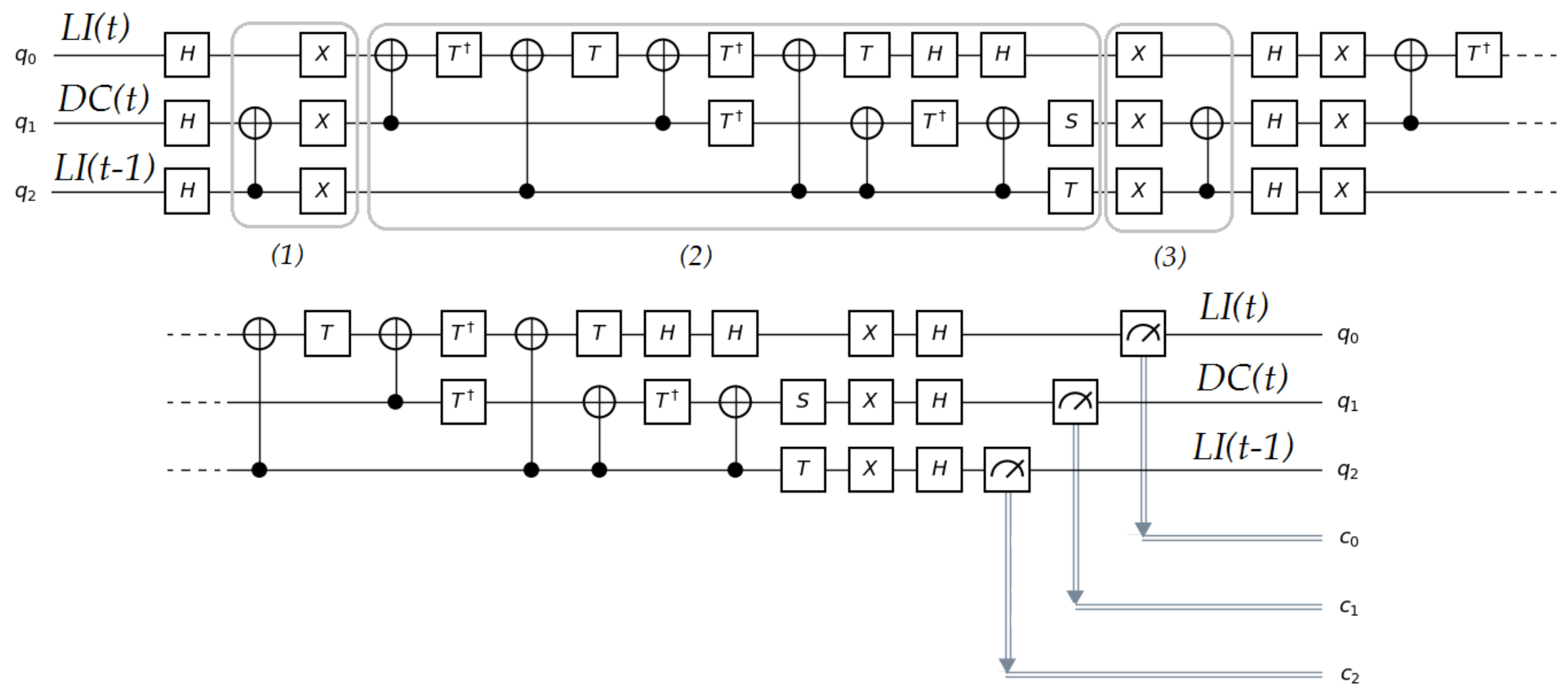

Figure 1: Quantum circuits implementing the rule $(\mathrm{LI}(t-1) \oplus \mathrm{DC}(t))' \cdot \mathrm{LI}(t-1)' . \mathrm{LI}(t)' = 1$ on the ibmqx4

If the generation continues (as opposed to requesting new candidates from the QC) then a second set of checks are done. If the returned solution for $\mathrm{LI}(t)$ is 1 (i.e. a large interval is required) then the generated white note interval will be limited to a maximum size of 8 (including both notes in the interval), otherwise it will be limited to a size of 3. The minimum size in both cases is 0, i.e. no pitch change. If the returned solution for $\mathrm{DC}(t)$ is 1 (i.e. a direction change is required) then a white note interval is generated between -8 and 8 (if a large interval is allowed) or between -3 and 3 (if a large interval is not allowed). Otherwise the generated interval will be in the same direction as the previous generated interval. The limiting of large intervals to white note size 8 is arbitrary, allowing for octave jumps at the most.

It can be seen that this implementation of Grover's algorithm in qgMuse is not generate-and-test. It would be more accurately described as solve-and-generate. The future quantum increase in speed would come from solving the Boolean equation. The process of turning the solution into pitches is implemented classically. This requires thoughtful designing of the sub-rules. The easy case is when the returned solution cannot be satisfied by the music generated thus far - in which case Grover's algorithm can be called again. Suppose a solution is found, and it involves the QC returning values for a large number of variables (say, 30) based on the Grover algorithm output. Then depending on the nature of these variables, a classical rule-based search may need to be run, to find solutions that give the allowed values for the variables. This could involve constraint satisfaction techniques. Thus the qgMuse approach does not replace constraint-based approaches, it provides a possible method for redesigning such approaches to speed up the parts that could involve generate-and-search sub-processes. For example, suppose the qgMuse quantum element returns 1, 1 and 0 for $\mathrm{LI(t)}, \mathrm{LI(t-1)}$ and $\mathrm{DC(t)}$. This could be viewed as setting up another set of rules:

$$(\mathrm{LI(t)} = 1) \cdot (\mathrm{LI(t-1)} = 1) \cdot (\mathrm{DC(t)} = 0) = 1 \tag{26}$$

or

$$\mathrm{LI(t)} \cdot \mathrm{LI(t-1)} \cdot \mathrm{DC(t)}' = 1 \tag{27}$$

This is simpler to solve than the full combined rule defined earlier in Equation (25).

It is so simple that it is easily implementable programatically, as done in this version of qgMuse. Also - as was already mentioned - the quantum approach provides "softness" for free. It is worth mentioning at this point that the softness is not as controllable as that available in some soft rule-based systems. A classical soft rule-based system could use defined probability distributions to ensure that the softness is controlled. Whereas the softness produced by Grover is a function of the quantum indeterminacy and errors in quantum hardware - a less controllable combination. But in terms of speed, random number generation in Grover's algorithm is "instant" - it does not require probability distribution or pseudorandom algorithms - the quantum mechanics provides true randomness instantly.

**5.2 Results**

Two runs of qgMuse of eight 4/4 bars each on the ibmqx4 quantum computer are shown in Figures 2 and 3. Examining limited melody samples does not fully characterise the system, however for clarity Table 2 gives the number of white notes spanned by each interval. There is a key point to be clarified here. The quantum part of qgMuse treats the large interval sub-rule in Equation (23) symmetrically. In other words if there is a small interval, there should be no direction change. The classical part of qgMuse , as described in the last section, uses the symmetrical solution of Equation (23) in an asymmetric way. The allowed interval sizes are defined based on the previous direction change, but current direction change is not restricted based on the previous interval size. Hence a small interval followed by a direction change is allowed in qgMuse as a whole, even though it is not a solution to Equation (25). It should also be noted that a 0 interval is not classed as being in any particular direction in the classical part of qgMuse, and so cannot break a direction rule.

Table 2. Number of white notes spanned by each melodic interval in Figures 3 & 4.

| Index | 1 | 2 | 3 | 4 | 5 | 6 | 7 | 8 | 9 | 10 | 11 | 12 | 13 | 14 | 15 | 16 | 17 |
|---|---|---|---|---|---|---|---|---|---|---|---|---|---|---|---|---|---|
| Melody A | 0 | -2 | -2 | -2 | -2 | 2 | 2 | 3 | 3 | -3 | -6 | 0 | -3 | 0 | 9 | 3 | -2 |
| | | | | | | | | | | | | | | | | | |
| Melody B | 3 | 3 | -3 | -2 | 3 | -3 | 3 | 2 | 8 | 0 | -3 | -2 | -2 | 2 | -2 | 0 | -2 |

| Index | 18 | 19 | 20 | 21 | 22 | 23 | 24 | 25 | 26 | 27 | 28 | 29 | 30 | 31 |
|---|---|---|---|---|---|---|---|---|---|---|---|---|---|---|
| Melody A | 0 | 2 | 4 | 3 | -7 | 2 | 6 | -3 | 0 | 8 | 3 | 8 | 0 | -3 |
| | | | | | | | | | | | | | | |
| Melody B | -2 | -8 | 0 | -2 | -2 | 2 | 7 | 3 | 2 | 0 | 5 | 2 | 3 | -2 |

Table 2 shows the number of white notes spanned by each melodic interval in Figures 2 and 3. From Table 2 it can be seen that qgMuse breaks the "no large interval without a following direction change" a number of times. For Melody A (Figure 2) the rule is clearly broken at interval indices 15, 20 and 27 (the rule is clearly followed at interval indices 22 and 24). For melody B (Figure 3) the rule is broken for interval indices 24 and 28. Thus in these small samples, the rule is clearly broken 5 times and is clearly followed twice. To give further insight, Figure 4 shows a plot of the output of the Grover algorithm in Figure 1 sampled 4096 times. IBM's Qiskit API returns the list of qubits as a single binary number. So in Figure 4 on the x-axis, 000-binary indicates all qubits are observed as 0. Similarly 111-binary indicates all qubits are observed as 1, and binary 101 means $q_1$ is observed as 0, but $q_0$ and $q_2$ are observed as 1. The solution of equation (25) can be seen by eye to be: all variables are 0. In the quantum world, this means that the most likely result of a measurement is all variables as 0. Figure 4 shows that the ibmqx4 returns all 0 approximately 52% of the time. The other 48% of the time, the incorrect values will be returned. It was initially desired that for soft rules, correct values be returned only most of the time. However a likelihood of 52% for a correct value, and 48% for a random one, is probably lower than desired for most soft rules. However, running Grover's algorithm 4096 times on the IBM online simulator gives Figure 5. This shows that the correct results are returned approximated 77% of the time, with a roughly equal spread across the incorrect results - totalling up to around 23%. This demonstrates that of the approximately 48% of incorrect results in the ibmqx4, around half of them are due to quantum hardware errors, rather than designed quantum effects. 23% error in soft rules seems a more reasonable figure - and is approximately what the ibmqx4 would presumably give if it was a perfectly engineered quantum computer (i.e. more like the simulator).

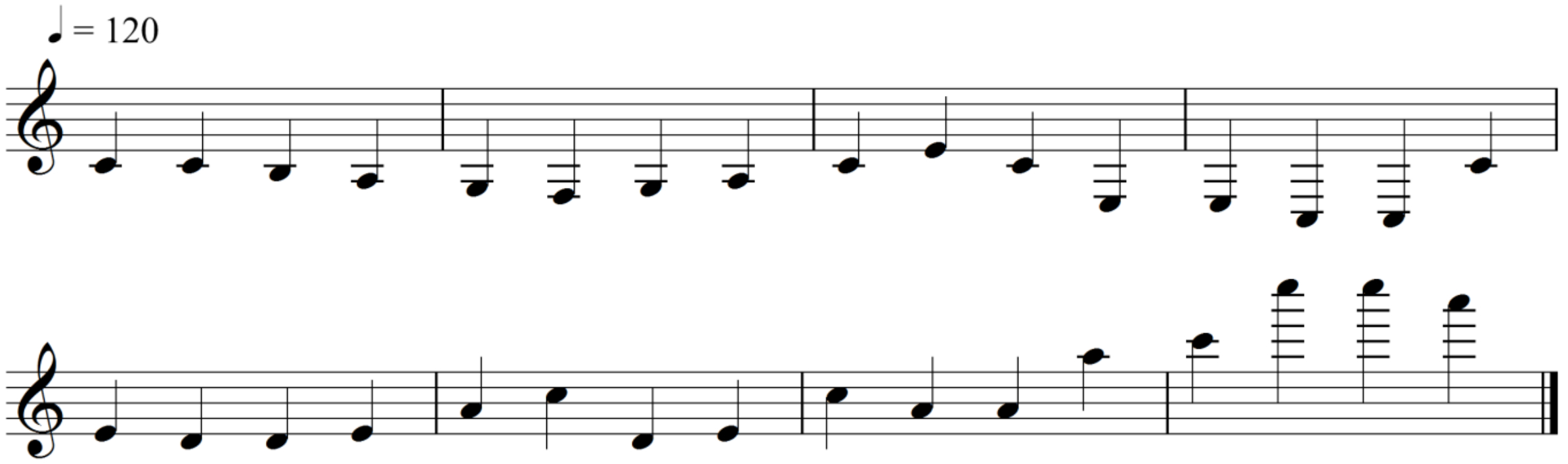

Figure 2:  8 bars of qgMuse running on the IBM quantum hardware - Melody A

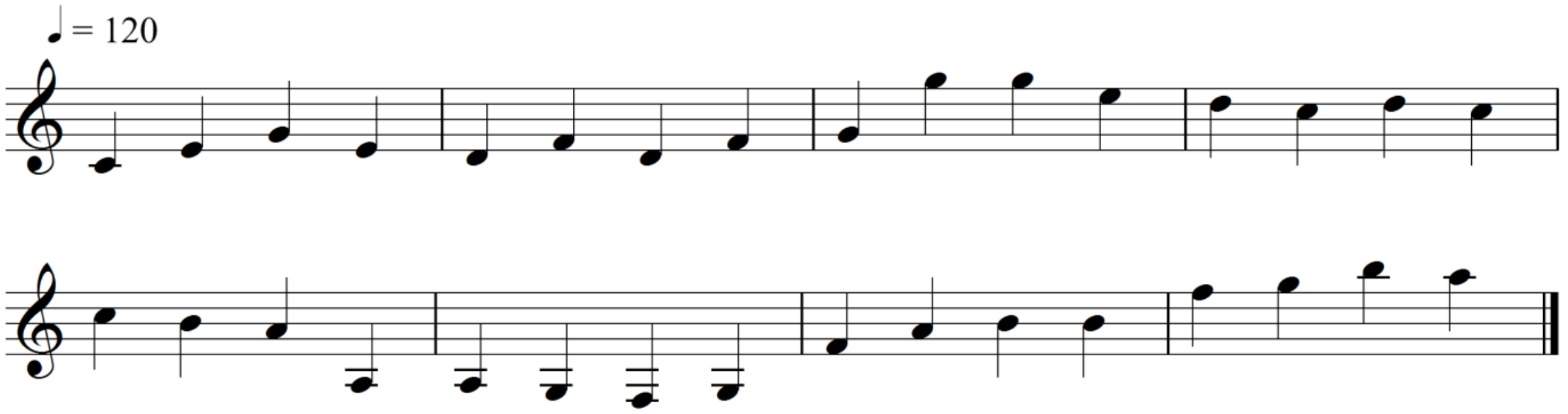

Figure 3: 8 bars of qgMuse running on the IBM quantum hardware - Melody B

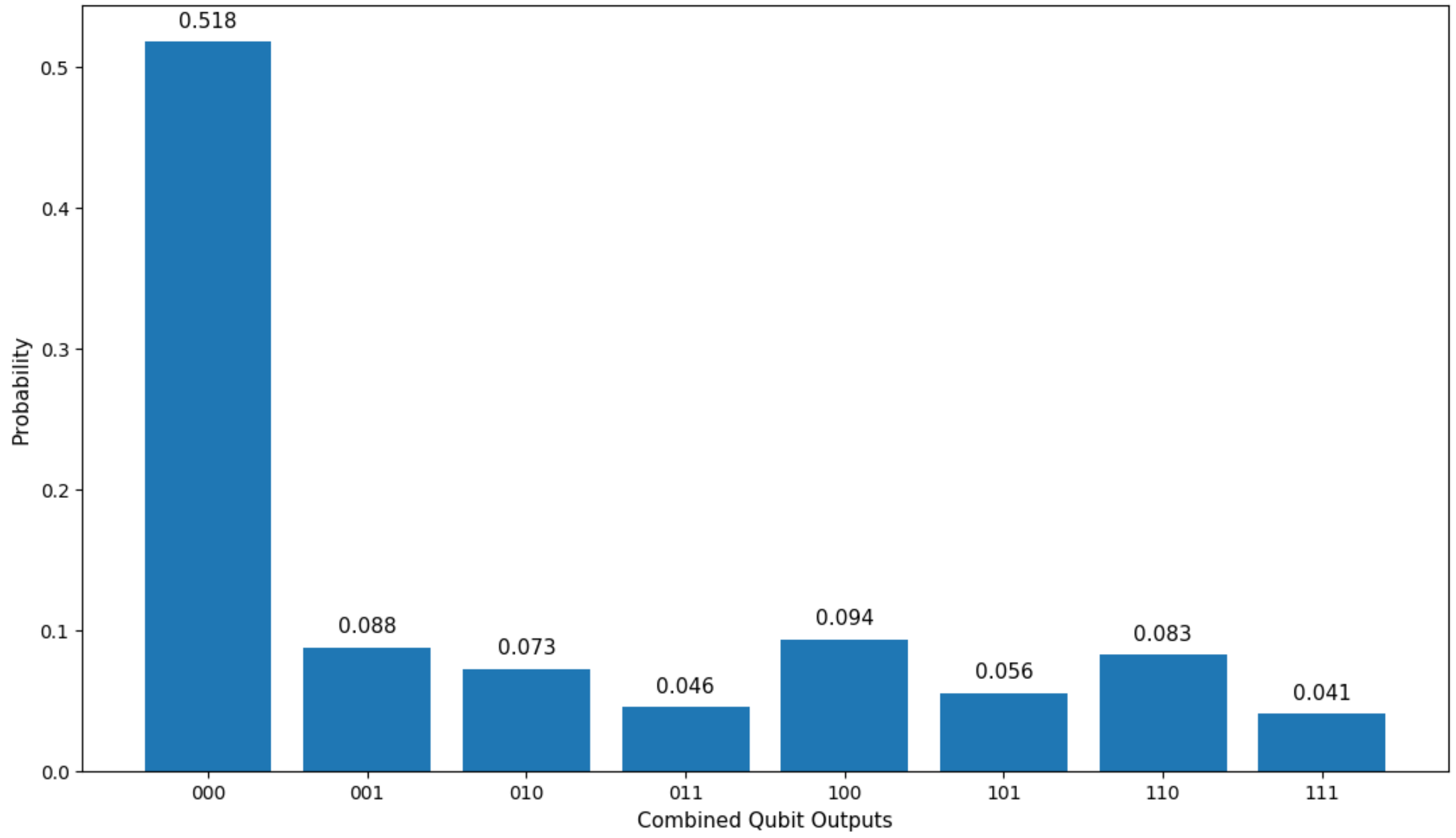

Figure 4: Grover rule outputs shown for the IBM quantum hardware ibmqx4 (refer to main text for details).

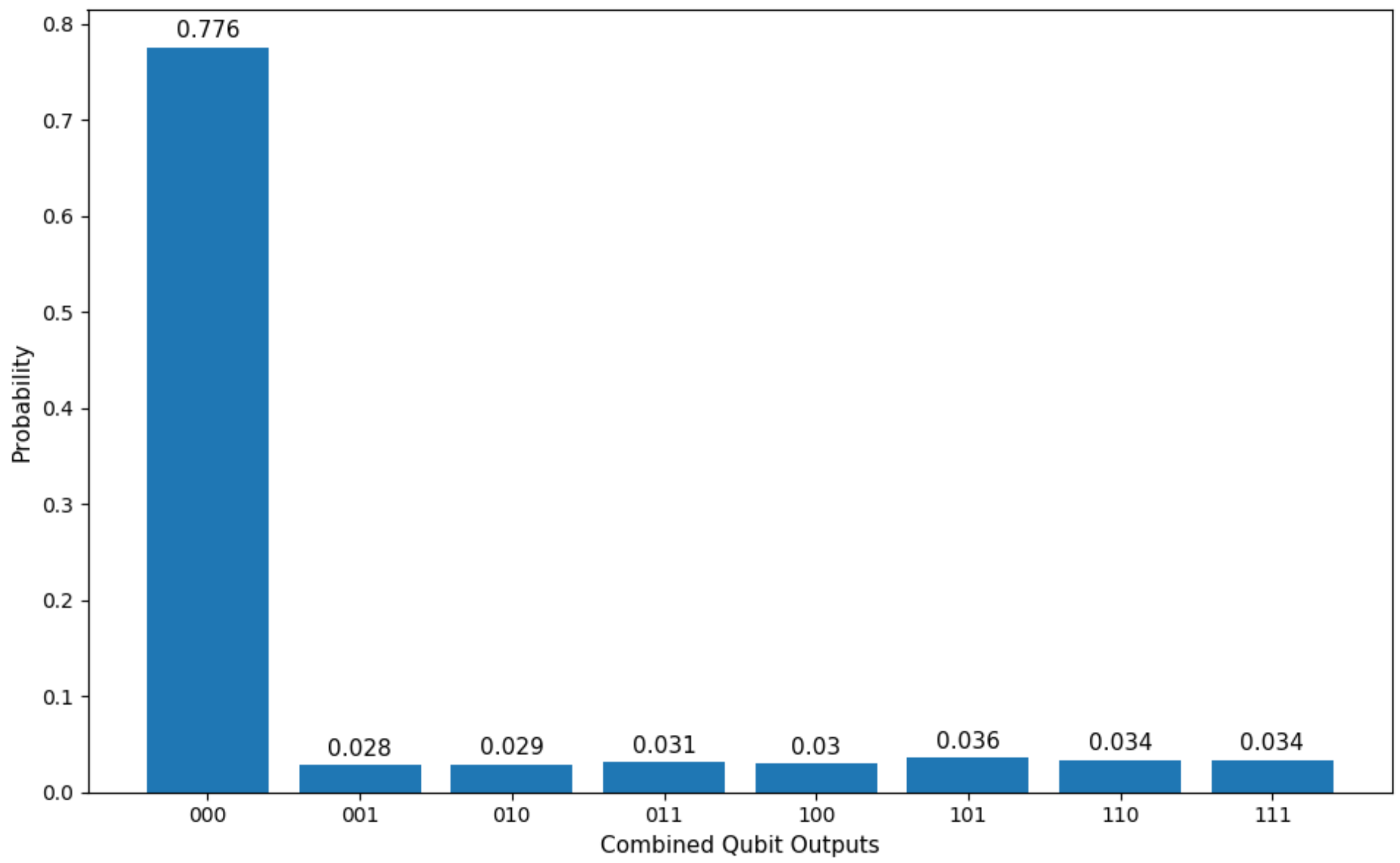

Figure 5: Grover rule outputs shown for the IBM online simulator (refer to main text for details).

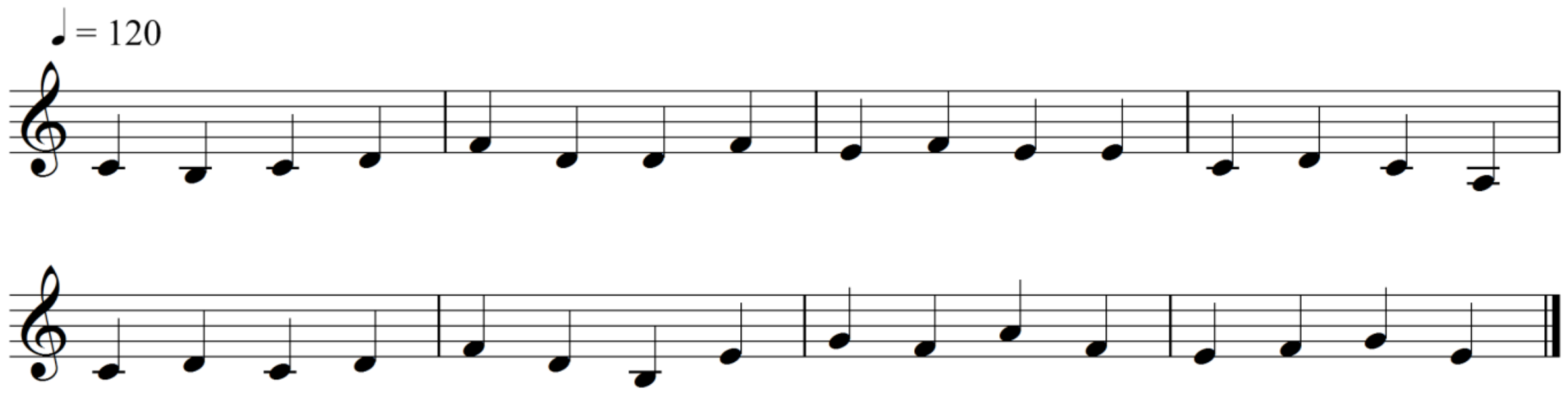

Figure 6: 8 bars of qgMuse running on the IBM quantum simulator - Melody C

Thus given that Figures 2 and 3 should only exhibit 52% correct behaviour, it is not feasible to use them to evaluate the statistical results. Generating larger and larger tunes to increase statistical clarity is pointless - Figure 4 (together with the deterministic behaviour of the classical computer code part of qgMuse) fully characterises qgMuse, and Figures 2 and 3 simply prove the feasibility of a musical output. However, the Figure 6 score of the online quantum simulator running qgMuse does allow some judgement by eye of its musical output (in terms of rule-following). All of the intervals except one in Figure 6 are non-large / small intervals (i.e., major thirds or smaller) - showing that one sub-rule (equation (24)) acting more consistently than in the hardware results. Given this, it is no surprise that the sub-rule of direction changes coinciding with a previous large interval (equation (23)) is not obviously visible in Figure 6. On the one occasion there is a large interval (bar 6, B -> E), it is not followed by the direction change required by equation (23).

It has been observed in the past that without some independent harmonic context, algorithmic melodies are difficult to evaluate [28]. Figure 7 shows the run of 8 bars output from Figure 2, this time with an independently generated harmonic context. The chords are generated using qHarmony [6] - an algorithm for generating simple white note harmonies without any temporal context. It takes as input one or two white notes, and outputs a white note chord to play with the notes. In this case two notes - the first and third note of each bar - are sent to qHarmony to select the chord for the whole bar. It does not use context of the previous or following chords. It is interesting to note that qHarmony is also run here on quantum hardware - it was in fact the first quantum hardware music algorithm - but on a very different system to the ibmqx4. It is a D-Wave 2X adiabatic quantum computer, referred to in the introduction to this paper as a quantum annealer. The algorithm is beyond the scope of this paper, but more information about qHarmony and adiabatic quantum computing can be found here [6]. qHarmony has previously been utilized in both live performance [27] and offline testing. There is one slight artistic license taken with the harmonies in Figure 7. When

qHarmony returns a harmonisation - it actually returns a selection of harmonisation solutions, together with their "energy level" for each. The energy level is an output parameter of the D-Wave 2X when it returns a solution to a problem. In theory, the lower the energy level, the better the solution. Bars 7 and 8 both returned the same solution chords at the lowest energy level. To create a small sense of movement between the penultimate bar and the final bar, the second lowest energy (energy -54) solution for Bar 7 was selected instead of the lowest energy solution (energy -56).

The tunes sound reasonably presentable, including the harmonized version. Some of this is due to the fact of the implicit rule of qgMuse - using only piano white notes. However a high level of musical quality is not expected from the limited sub-rule-set size feasible on current hardware QCs.

The results of qgMuse above support the idea of the non-deterministic nature of current quantum machines being a potential "feature rather that a bug" - in terms of implementing soft rules. Furthermore, Figures 4 to 6 are consistent with the idea that the rules are being followed the majority of the time. Given the usage of a quantum computer with a similar level of hardware error (in the final outputs), but with a much larger number of qubits, then qgMuse can be scaled in a simple way. The number of sub-rules can be increased. This can be done with conjunction or disjunction - i.e. by ANDing the rules (as described earlier) or ORing them, or a combination (a quantum OR can be built using X gates and ccx gates). Then each Boolean input can be assigned to a qubit.

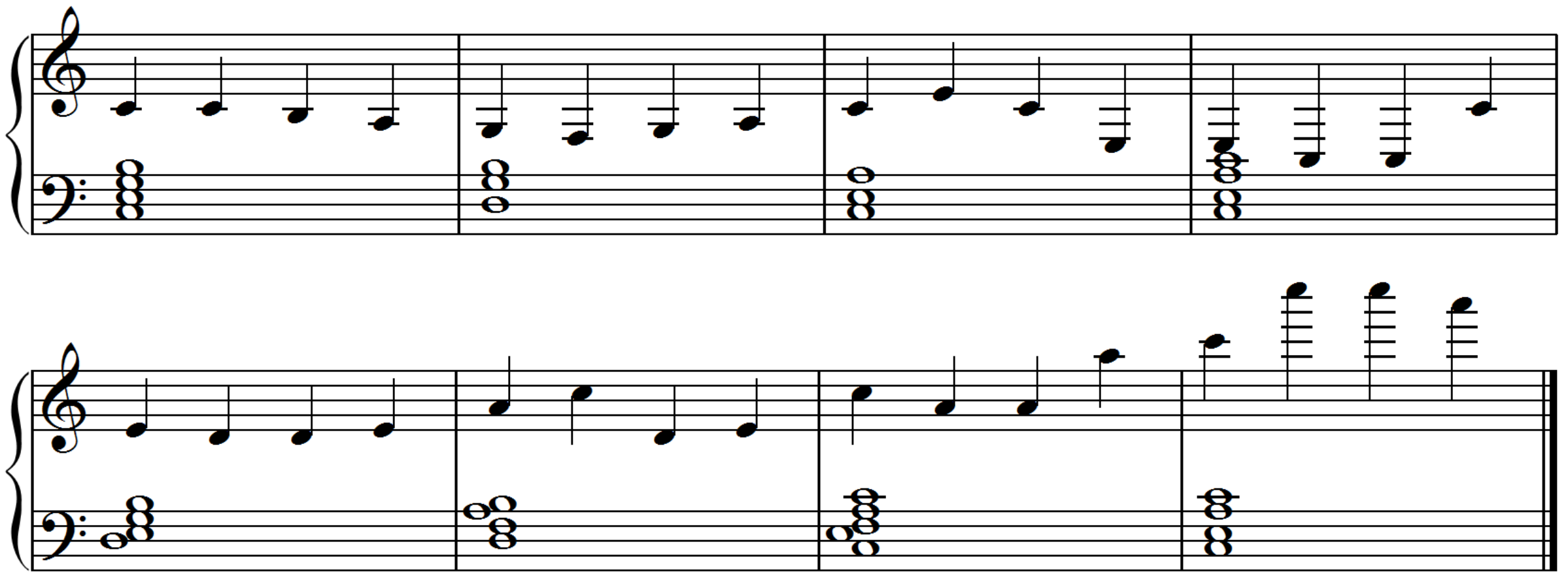

Figure 7: The melody in Figure 5 - Melody A, harmonised by an independent harmony algorithm, qHarmony, on a D-Wave 2X adiabatic quantum computer.

Note that the proviso above about the level of error in the final outputs is key. Just increasing the number of qubits and the size of the Boolean functions, at the current level of gate error, will create an essentially random output. The input / output error refers to this: given a certain input - and given the resulting output - how far away are the outputs from the outputs of a perfect simulator? To achieve the same level of error as seen in Figure 4 for 100 gates is beyond current quantum technology. However if the gate errors can be kept small enough, so that perhaps 500 gates can be combined with the same level of input / output error as seen in Figure 4, then more interesting musical problems become possible using Grover. Given this level of error, exactly the same methods can be used as are presented in this paper, the main challenge becoming - how to implement Oracles for the various musical rules.

Of course, at the current level of simplicity of qgMuse required by quantum technology, the whole rule-based process in qgMuse process could have been done trivially. Rather than running Grover's algorithm, the results of the randomly generated notes and of calculating LI and DC could have been inserted into the classical Boolean equations. If they failed to fulfil those equations, then 75% of the time (say, to allow for softness) the proposed interval could have been ignored and a new one generated. This approach works fine until the number of equations and inputs grows large. Then various heuristics and optimisation methodologies need to be introduced. Gradually as the number of rules and inputs grows, the result becomes very slow to generate on a classical system. At this point the problem moves into the domain where a Grover-assisted algorithm using the ideas of qgMuse, on a sufficiently powerful future quantum computer, would have an advantage.

The reality is that a future top-down rule-based system for quantum computer music will probably be a hybrid between constraint programming and quantum computer unstructured search. The quantum part will be the Grover's algorithm used in this paper, but implemented in more error-tolerant ways and utilizing quantum error reduction algorithms [44]. But whatever the precise balance and approach used, the basic structure of a Grover solving a Boolean rule will remain the same, as described in this paper. It can in fact be proved that Grover's algorithm is optimal for unstructured search [45].

## 6. CONCLUSIONS

In evaluating the design and results of qgMuse, context is key. For example, qgMuse is slower than a fully classical version. The quantum advantage of Grover is superseded by the simplicity of the rules, and the small number of inputs. A classical version will be faster, as accessing the small number of quantum computers sometimes involves a queue time, and obviously running the simulator is slower than the classical equivalent. Similarly, from a musical point of view, there are only two simplistic rules being used, which are applied to a musical context of only one previous note and a small number of music features.

Thus the evaluation context is as follows: (a) Shor's algorithm is currently only factoring numbers of the order of 15; (b) qgMuse is scalable as a concept; and (c) qgMuse has a natural soft non-deterministic nature. Examining the second point, which is overwhelmingly the most important: suppose there is a conjunction of 150 Boolean musical sub-rules with 30 inputs, similar to that discussed earlier. Finding solutions to that conjunction by classical unstructured search would be unfeasible, and may never become feasible given the limitations of Moore's law [40]. However quantum computing is moving forward at a significant rate in 3 key areas: number of qubits, stability of processing, and error reduction methods. When the 16 qubit IBM Melbourne was released, it came rapidly on the heels of the 5 qubit ibmqx4. Other companies claim to have QCs with greater than 60 qubits, but many are not publicly available to use yet. The instability of hardware quantum computers is not a new issue, and there have been many years of research [46] into how to reduce the susceptibility of quantum computing to errors such as those encountered in this paper, including on the ibmqx4 itself [44]. In the next couple of years, someone may publish an error tolerant version of the 3-input Grover for the IBM q 16 Melbourne. It is not unfeasible that in 4 years the IBM q 16 Melbourne will be superseded by a much more stable q 30, on which Grover's algorithm, utilizing error reduction, could solve 10-20 variable problems. There are researchers who say that the demonstrating of quantum supremacy over classical computers in hardware for certain algorithms is only a matter of years away [47]. It is not yet clear, however, how this might relate to time scales for a similar demonstration with Grover’s algorithm.

The ibmqx4 has fidelity of 99.7% for one qubit gates and 95.8% for two qubit gates [48]. These are measures of the accuracy of the single input and two input gates. Two-input gates such as cx are vital, but are much harder to implement than single input gates such as x. The 16 qubit Melbourne has 99.7% and 92.8% for these values - more error prone than the ibmqx4 (in fact the experiments in this paper were first run on the Melbourne, but then discarded due to impractical error rates). Google have built a 72-qubit quantum computer and report accuracies of 99.9% and 99.4%. Recently IonQ announced a 79 qubit machine and report rates of 99.97% and 99.3% [49]. While all of this is happening, it is expected that the research community will develop a strong subset of research looking at how to implement Boolean solving problems on quantum computers. IBM have released a machine learning kit for their computers, but they have not tested it running Grover on Boolean equations on hardware.

The Grover's algorithm approach taken in qgMuse is fairly simplistic and linear. CHORAL and its progeny utilize many ingenious methods to speed up the search for solutions. Quantum-enhanced versions of these methods are sure to emerge. The computer arts community can utilize such advantages.

The computer arts world can also contribute to the quantum computing research world in its crossover from unconventionality to conventionality. For example, computer music can provide an environment for utilizing quantum algorithms that - because of music's relatively lightly structured nature - will help programmers who are non-quantum physicists and non-mathematicians to gain greater practical insight. Machine learning / statistical analysis for language processing or molecular pattern recognition are highly constricted problems leading to extremely technical results, whose results are meaningless to most non-experts. Music and sound allow for freer output structures - that can still produce a sense of meaning and pattern, with relatively light-touch algorithms. Hence a quantum computer musician may write a paper explaining their work and its results in a far more accessible way than a quantum computational chemist.

## 7. ACKNOWLEDGEMENTS

Thanks to Vlatko Vedral of Wolfson College, Oxford and the Centre for Quantum Technologies, Singapore; and to Miklos Santha, Director of Research at the Research Institute on the Foundations of Computer Science (INRI) at Université Paris Diderot. Thanks also to Daniel Lidar, Director of the USC Center for Quantum Information Science and Technology, and Scientific Director of the USC-Lockheed Martin Center for Quantum Computing. Particular thanks to Yassine Hamoudi of INRI - whose optimization and error-checking of the Grover's algorithm part in qgMuse was fundamental to its success.